\documentclass{article}

\usepackage{microtype}
\usepackage{graphicx}
\usepackage{booktabs}
\usepackage{amsmath}
\usepackage{amssymb}
\usepackage{multirow}
\usepackage{xcolor}
\usepackage{hyperref}

\usepackage[accepted]{mlsys2025}
\makeatletter
\renewcommand{\Notice@String}{Preprint.}
\makeatother
\hypersetup{pdfauthor={Oguzhan Buyuksolak, Osman Erman Okman, Mehmet Gorkem Ulkar}, pdfsubject={Preprint}}

\usepackage{algorithm}
\usepackage{algorithmic}

\usepackage{iftex}
\ifXeTeX
  \usepackage{fontspec}
\fi

\mlsystitlerunning{ShiftQAT: Quantization-Aware Training for a Fixed-Function Edge Accelerator}

\begin{document}

\twocolumn[
\mlsystitle{ShiftQAT: Quantization-Aware Training for the Numerical Contract\\of a Fixed-Function Edge CNN Accelerator}

\begin{mlsysauthorlist}
\mlsysauthor{Oguzhan Buyuksolak}{adi}
\mlsysauthor{Osman Erman Okman}{adi}
\mlsysauthor{Mehmet Gorkem Ulkar}{adi}
\end{mlsysauthorlist}
\mlsysaffiliation{adi}{Analog Devices, Inc}
\mlsyscorrespondingauthor{Oguzhan Buyuksolak}{oguzhan.buyuksolak@analog.com}

\mlsyskeywords{Quantization-aware training, edge inference, CNN accelerator, power-of-two quantization, TinyML, MLSys}

\vskip 0.3in

\begin{abstract}
Fixed-function CNN accelerators make edge inference fast and energy-efficient, but they execute only networks that obey a precise numerical contract.
On the MAX7800x family, this contract requires power-of-two, per-tensor, symmetric scales, 8-bit activations and biases, integer-only arithmetic, and one statically configured shift per layer.
We present ShiftQAT, the quantization-aware training (QAT) flow we developed for this family, which produces networks satisfying the contract by construction. It is the default training path of ai8x-training, the family's open-source training toolchain.
ShiftQAT resolves residual, concatenation, and weight-sharing dependencies so that activation scales propagate statically and each layer's requantization reduces to one offline-configured shift.
It also trains weights and biases on a joint power-of-two grid that fits the 8-bit bias budget.
Among the compared methods, only ShiftQAT satisfies all of the contract's scaling constraints, including signed 8-bit activation storage. Even so, on ImageNet it matches the W8A8 accuracy drop of less constrained QAT methods and stays within 2.9 percentage points of FP32 at W4A8.
On MAX78002 silicon, ShiftQAT-trained MobileNetV2 and ResNet-18 classify ImageNet images in 14.6 and 8.4~ms at 5.1 and 2.8~mJ per inference.
\end{abstract}
]

\printAffiliationsAndNotice{}

\section{Introduction}
\label{sec:intro}

Deep neural networks are increasingly deployed on battery-powered edge devices, close to the sensors that produce their inputs.
General-purpose microcontrollers offer only a few megabytes of memory and limited arithmetic throughput. This has motivated software stacks for running quantized models on MCUs~\citep{lai_2018_cmsisnn, david_2021_tflm, lin_2020_mcunet}, as well as dedicated low-power CNN accelerators integrated alongside the MCU core~\citep{giordano_2022_aicas, huang_2024_energycharacterization}.
Quantization is the enabling step for both, because it shrinks the memory footprint and lets inference run in fixed-point arithmetic~\citep{gholami2022survey, nagel_2021_whitepaper}.

The quantization literature usually treats hardware friendliness as a spectrum of desirable properties, such as uniform, symmetric, power-of-two, and per-tensor scaling. Methods are then compared by how much accuracy they recover at a given bit-width.
Fixed-function accelerators turn these properties into hard requirements.
We call the complete set of arithmetic rules that an accelerator's datapath implements its numerical contract. A network either satisfies this contract or cannot be compiled for the device at all.
On Analog Devices' MAX7800x family~\citep{max78002}, the contract requires power-of-two, per-tensor, symmetric scales, 8-bit storage for activations and biases, and integer-only inference. Each layer has a single shift that is configured offline and cannot rescale dynamically where branches meet (Section~\ref{sec:contract}).
Prior hardware-friendly methods each satisfy part of this contract, but none of those we compare satisfies all of it (Tables~\ref{tab:w8a8} and~\ref{tab:w4a8}).

This paper presents ShiftQAT, a QAT flow that produces networks satisfying the full contract by construction.
We developed ShiftQAT at Analog Devices as the default quantization-aware training path of ai8x-training, the open-source training toolchain for the MAX7800x family.
This paper documents its design, the datapath constraint behind each of its mechanisms, and its evaluation on ImageNet and on MAX78002 silicon.
Each of its mechanisms answers a specific constraint of the datapath, and two of them carry the design.
The first is static scale propagation.
Because the datapath cannot rescale dynamically, ShiftQAT resolves all rescaling before deployment. It folds each layer's activation scale forward into the next layer's bias and shift, so that intermediate layers never recover real-valued scale.
This requires positively homogeneous activations and a scale-consistent graph. ShiftQAT therefore resolves every residual addition, concatenation, and shared-weight layer so that propagation remains exact on real topologies.
The second is a joint power-of-two weight/bias scale, re-estimated at every training iteration.
Once biases have absorbed upstream activation scales and must still fit in 8 bits, weights and biases compete for the same power-of-two grid. Training on that grid lets the network adapt to it.
The per-layer shift that the hardware applies is computed by training itself. A trained checkpoint therefore maps deterministically to integer weights, 8-bit biases, and shift values, with no separate calibration step.

Our contributions are:
\begin{itemize}
\item We formalize the numerical contract of a fixed-function edge CNN accelerator as seven datapath constraints (Table~\ref{tab:contract}). None of the prior methods we compare satisfies all of them.
\item We present ShiftQAT, which satisfies the full contract through topology dependency resolution, static scale propagation, and joint power-of-two weight/bias scaling, specified end-to-end in Algorithm~\ref{alg:shiftqat}.
\item Under the contract's scaling constraints, ShiftQAT matches the ImageNet accuracy drop of less constrained QAT methods at W8A8 and stays within 2.9~pp of FP32 at W4A8 on MobileNetV2 and ResNet-18. We quantify the gap to QAT methods that relax the contract at 4 bits.
\item We deploy ShiftQAT-trained MobileNetV2 and ResNet-18 end-to-end on MAX78002 silicon and report measured latency, energy, and memory for ImageNet-scale inference on a microcontroller-class device.
\item We distill five lessons for building training flows for fixed-function accelerators (Section~\ref{sec:lessons}). The complete flow is available as open source in ai8x-training.
\end{itemize}

\textbf{Code availability.}
The code is published at ai8x-training, \url{https://github.com/analogdevicesinc/ai8x-training}.

\section{Background and Related Work}
\label{sec:related}

\begin{figure*}[t]
\centering
\includegraphics[width=0.95\textwidth]{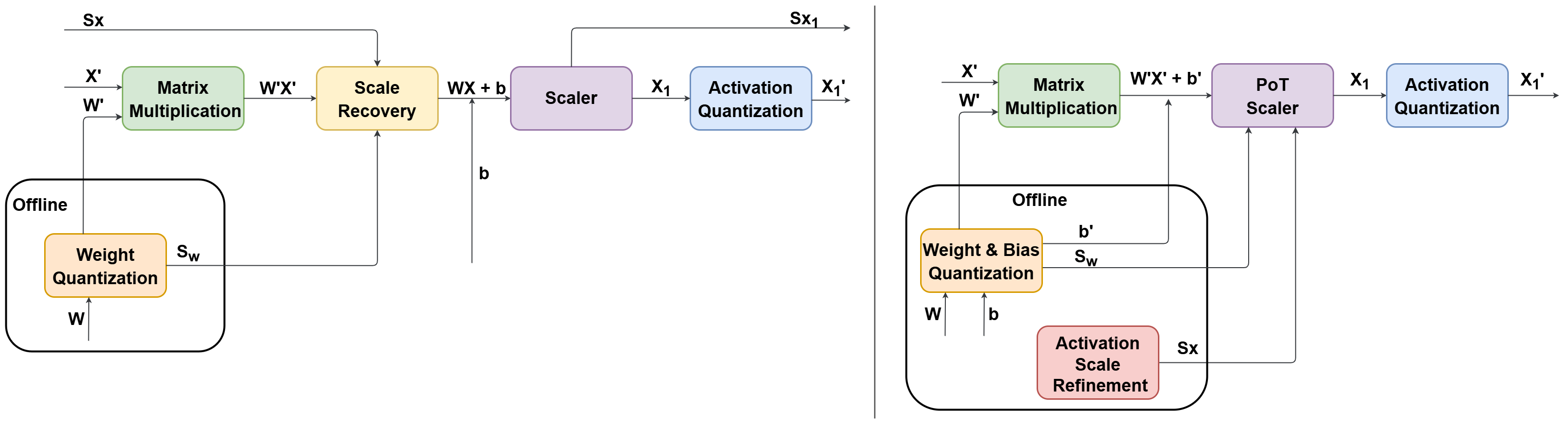}
\caption{Per-layer inference datapaths. (a) Per-layer scale recovery as in integer-only QAT~\citep{jacob_2018_CVPR, esser_2020_lsq}. The product $W'X'$ is rescaled by $S_w$ and $S_x$ before a full-precision bias is added, and the result is then requantized. (b) The structure the MAX7800x contract requires, as produced by ShiftQAT. Weights and biases share one power-of-two scale $S_w$, and activation scales are resolved and propagated offline. The only runtime rescaling is a single power-of-two shift per layer (the PoT scaler, Eq.~\ref{eq:pot}).}
\label{fig:datapath}
\end{figure*}

\textbf{Integer-only inference and dyadic requantization.}
\citet{jacob_2018_CVPR} formalized integer-only inference.
Each layer's real-valued output multiplier is decomposed offline into a fixed-point mantissa and a power-of-two shift, and biases are stored as 32-bit integers at the accumulator's scale.
HAWQ-V3~\citep{yao_2021_hawqv3} restricts these multipliers to dyadic numbers, and TensorRT~\citep{migacz_2017} and CMSIS-NN~\citep{lai_2018_cmsisnn} provide calibrated or Q-format fixed-point kernels for GPUs and Cortex-M CPUs.
Figure~\ref{fig:datapath}a shows this per-layer recovery structure.
On processors with a general integer multiplier it is inexpensive.
A fixed-function accelerator changes where the rescaling can happen. It can only be a statically configured power-of-two shift per layer, never a multiplication by an arbitrary mantissa.

\textbf{Moving scales across layers.}
An activation $f$ is positively homogeneous if $f(sz) = s\,f(z)$ for all $s>0$, as holds for ReLU and Leaky ReLU. For such activations, a per-channel or per-tensor scale can be moved through the activation into the next layer's weights.
Data-free quantization~\citep{nagel_2019_dfq} exploits this property for cross-layer weight equalization.
ShiftQAT uses the same identity for a different purpose. It defers every activation scale to the network output, so that intermediate layers never recover real-valued scale.
It adds the topology handling (Section~\ref{sec:deps}) and the bias treatment (Section~\ref{sec:wb}) needed to make this hold under an 8-bit bias budget on real CNNs.

\textbf{Hardware-friendly quantizers.}
PACT~\citep{choi_2018_pact}, FAQ~\citep{mckinstry_2019_faq}, LSQ~\citep{esser_2020_lsq}, and differentiable range/step-size methods~\citep{uhlich_2020_clr} learn clipping ranges or step sizes during QAT.
TQT~\citep{TQT_2019} trains power-of-two per-tensor thresholds, and APoT~\citep{li_2020_apot} uses sums of powers of two as non-uniform levels. \citet{vandersteegen_2021_int4shift} combine per-layer power-of-two scales with 4-bit weights, optionally through a lookup table that maps 4-bit codes to non-uniform 8-bit weights.
Post-training methods such as AdaRound~\citep{markus2020adaround}, BRECQ~\citep{li2021brecq}, HPTQ~\citep{habi_2021_hptq}, and EPTQ~\citep{gordon_2024_eptq} avoid retraining.
The strongest of them rely on per-channel scales, which the MAX7800x datapath does not provide.
HPTQ is closest in spirit to ShiftQAT. It targets uniform, symmetric, power-of-two quantization, and ShiftQAT adopts its activation-threshold search (Section~\ref{sec:actscale}).
Lower-bit and binary/ternary schemes~\citep{hubara_courbariauc_nips2016, zhou2018dorefanettraininglowbitwidth, zhang_2018_eccv_lqnets, jung_2019_cvpr, yang_2021_iclr_bsq} pursue compression beyond the 8/4-bit regime studied here.

\textbf{Edge CNN accelerators.}
The MAX78000 and MAX78002 integrate a CNN accelerator with 1/2/4/8-bit weights and 8-bit activations alongside a Cortex-M4 core.
Independent studies have characterized their latency and energy on small vision and audio workloads~\citep{BED_acm_2022, giordano_2022_aicas, moosmann_2023_tinyssimoyolo, huang_2024_energycharacterization, dabbous_2025_tinyml_max78000}.
This paper describes the training side, namely the QAT flow of the family's own toolchain, together with ImageNet-scale deployments on it.

\section{The Accelerator's Numerical Contract}
\label{sec:contract}

Table~\ref{tab:contract} lists the constraints a network must satisfy to run on the MAX7800x CNN accelerator, together with the ShiftQAT mechanism each constraint forces.
Appendix~\ref{app:hwfq} expands on each property.

\begin{table}[t]
\centering
\caption{The numerical contract of the MAX7800x CNN accelerator, as specified in the public MAX7800x documentation, and the ShiftQAT mechanism each constraint requires. Constraints C1--C7 are properties of the datapath, not design preferences.}
\label{tab:contract}
\small
\setlength{\tabcolsep}{3pt}
\begin{tabular}{@{}p{0.08\columnwidth}p{0.47\columnwidth}p{0.38\columnwidth}@{}}
\toprule
& \textbf{Constraint} & \textbf{Required mechanism} \\
\midrule
C1 & Signed fixed-point (Q7) number format; uniform, symmetric, no zero-point & Symmetric quantizers everywhere (\S\ref{sec:actscale}, \S\ref{sec:wb}) \\
C2 & Rescaling only by $2^{k}$, $k\in[-15,15]$, applied to the 32-bit accumulator & PoT scales for $S_x$, $S_w$ (\S\ref{sec:actscale}, \S\ref{sec:wb}) \\
C3 & One output shift per layer (per-tensor) & Per-tensor $S_x$, $S_w$ \\
C4 & Biases stored in 8 bits; weights in 8/4/2/1 bits & Joint weight/bias scale (\S\ref{sec:wb}) \\
C5 & Activations stored as signed 8-bit; ReLU outputs clipped to $[0,127]$ & Signed symmetric activation quantization \\
C6 & No floating point in the inference path & All scales folded offline (\S\ref{sec:prop}) \\
C7 & Shift configured statically; element-wise operations combine inputs without per-input rescaling & Dependency resolution + propagation (\S\ref{sec:deps}, \S\ref{sec:prop}) \\
\bottomrule
\end{tabular}
\end{table}

\textbf{Why generic QAT output does not compile.}
A network produced by standard integer-only QAT~\citep{jacob_2018_CVPR} typically violates C2 and C4 (non-power-of-two multipliers, 32-bit biases), and often C1 (asymmetric activations).
Per-channel PTQ methods~\citep{li2021brecq, gordon_2024_eptq} additionally violate C3.
TQT~\citep{TQT_2019} comes closest. It uses power-of-two, per-tensor scales and merges the input scales of element-wise additions and concatenations, which addresses C7. However, it quantizes biases to 16 bits and switches to unsigned activations after ReLU, which violates C4 and C5. HPTQ~\citep{habi_2021_hptq} relies on per-channel weight scales, which violates C3.
The bias constraint C4 interacts with C7. Once activation scales are folded forward (C6, C7), biases absorb those scales and must still fit in 8 bits.
Satisfying the constraints one at a time therefore does not satisfy them jointly, which is why ShiftQAT treats the contract as a whole.

\textbf{What the contract rules out.}
Two constraints remove accuracy levers that most published methods rely on.
C3 forgoes per-channel scales, which the strongest PTQ results depend on.
C5 means that post-ReLU tensors use only 7 of their 8 bits. The hardware clips ReLU outputs to $[0,127]$ in its signed 8-bit format, so half of the code space is unused for non-negative tensors.
ShiftQAT recovers accuracy within these limits through training, as Section~\ref{sec:acc} shows.

\section{ShiftQAT Mechanisms Required by the Contract}
\label{sec:method}

\begin{figure*}[t]
\centering
\includegraphics[width=0.72\textwidth]{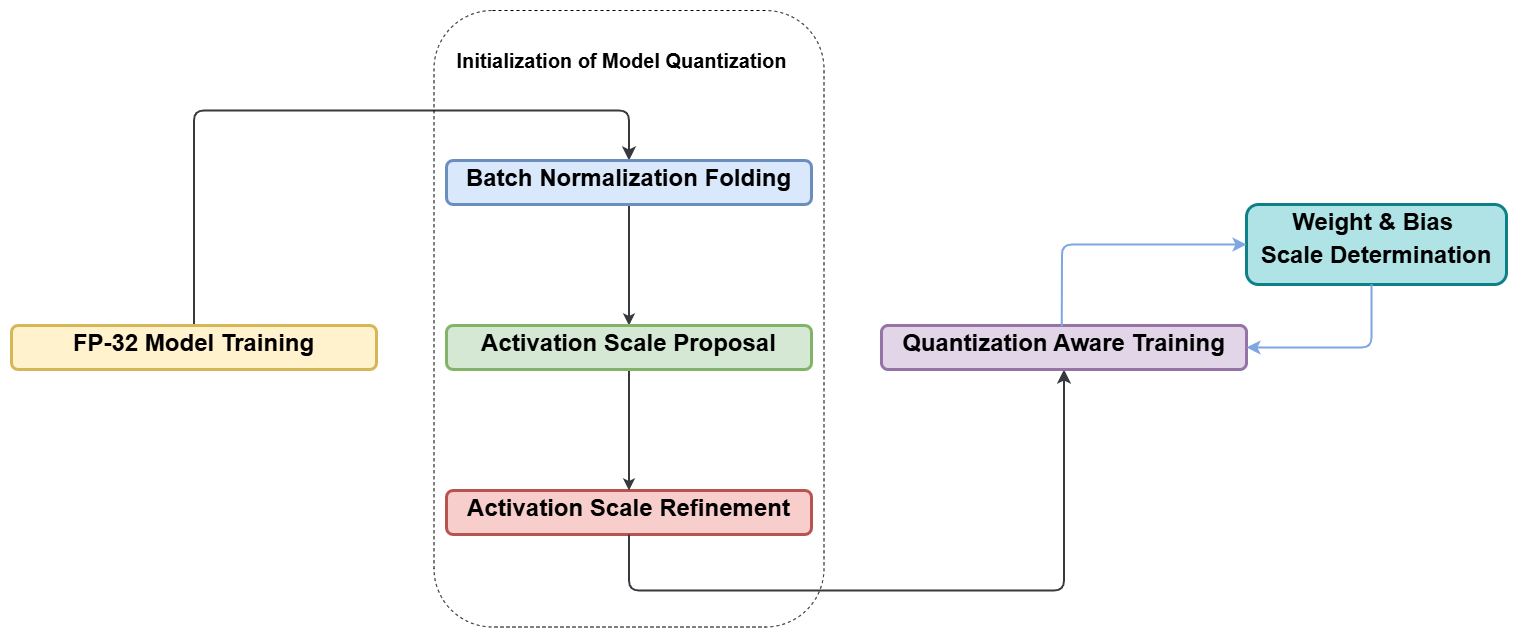}
\caption{The ShiftQAT flow. FP32 training is followed by offline initialization, which consists of BN folding, activation-scale proposal, and activation-scale refinement (dependency resolution and propagation). QAT then runs with per-iteration weight/bias scale determination.}
\label{fig:flow}
\end{figure*}

Figure~\ref{fig:flow} and Algorithm~\ref{alg:shiftqat} summarize the flow.
It starts from an FP32 checkpoint and runs an offline initialization consisting of batch-norm folding, activation-scale proposal, dependency resolution, and propagation. QAT follows, and the weight/bias scale is re-estimated at every iteration.
Throughout, $n_x$, $n_w$, and $n_b$ denote the bit-widths of activations, weights, and biases, and every scale $S=2^{s}$ is a power of two with integer exponent $s$.

\begin{algorithm}[t]
\caption{ShiftQAT, offline initialization and QAT loop}
\label{alg:shiftqat}
\small
\begin{algorithmic}[1]
\REQUIRE FP32 model $M$, training set $\mathcal{D}$, bit-widths $n_x, n_w, n_b$
\STATE \textbf{// Offline initialization}
\STATE Fold batch normalization into preceding layers
\FOR{each layer $\ell$}
  \STATE Collect activations on $\mathcal{D}$; drop z-score outliers
  \STATE $S_{\mathrm{nc}} \leftarrow$ smallest PoT covering the remaining range
  \STATE $S_{x,\ell} \leftarrow \arg\min$ MSE over $S_{\mathrm{nc}}/2^{i}$, $i=0,\dots,4$ (Eq.~\ref{eq:argmin})
\ENDFOR
\STATE Resolve dependencies (\S\ref{sec:deps}). Residual inputs take the add's scale, and concatenation and shared-weight inputs take the maximum scale
\FOR{each layer $\ell$ in topological order, with predecessor $p$}
  \STATE $b_\ell \leftarrow b_\ell / S_{x,p}$; \quad $S'_{x,\ell} \leftarrow S_{x,\ell}/S_{x,p}$ \hfill (Eq.~\ref{eq:prop})
\ENDFOR
\STATE \textbf{// Quantization-aware training}
\FOR{each training iteration}
  \FOR{each layer $\ell$}
    \STATE $\lambda_\ell \leftarrow$ $q$-quantile of $|\{W_\ell, b_\ell\}|$; \ $s_{w,\ell} \leftarrow \mathrm{clip}(\lceil \log_2 \lambda_\ell \rceil; -15, 15)$ (Eq.~\ref{eq:sw})
    \STATE Quantize $W_\ell$ ($n_w$ bits), $b_\ell$ ($n_b$ bits) with $S_{w,\ell} = 2^{s_{w,\ell}}$
    \STATE $k_\ell \leftarrow \mathrm{clip}(s_{w,\ell} - s'_{x,\ell}; -15, 15)$ \hfill (PoT scaler, Eq.~\ref{eq:pot})
  \ENDFOR
  \STATE Forward pass with fake quantization; backward pass with the STE; update $W, b$
\ENDFOR
\ENSURE Integer weights, 8-bit biases, one shift per layer
\end{algorithmic}
\end{algorithm}

\subsection{Batch-Normalization Folding}
Batch normalization~\citep{ioffe2015batch} is folded into the preceding convolution or linear layer before quantization, following \citet{jacob_2018_CVPR}.
This is required by C6. The accelerator has no normalization stage, so its affine transform must be absorbed into the weights and biases that the contract governs.

\subsection{Activation-Scale Proposal}
\label{sec:actscale}
For each layer we select a power-of-two activation scale (C2, C3) using the threshold search of HPTQ~\citep{habi_2021_hptq}.
Activation statistics are collected as 2048-bin histograms in one forward pass over the training set, and outliers are removed with a z-score threshold of 8.
With $S_x = 2^{k}$, $k\in\mathbb{Z}$, the step size is $\Delta_x = S_x/2^{n_x-1}$, and activations are quantized symmetrically (C1, C5):
\begin{equation}
X' = \mathrm{clip}\!\left(\left\lfloor X/\Delta_x \right\rceil;\, -2^{n_x-1},\, 2^{n_x-1}-1\right)\cdot\Delta_x ,
\end{equation}
where $\lfloor\cdot\rceil$ denotes rounding to the nearest integer.
The search starts from the smallest non-clipping power of two for the outlier-filtered activations,
\begin{equation}
S_{\mathrm{nc}} = 2^{\lceil \log_2 \max |X| \rceil}.
\end{equation}
It evaluates this scale and its first four halvings, and keeps the candidate that minimizes the quantization error weighted by the activation histogram:
\begin{equation}
S_x^{*} = \arg\min_{S_x \in \{S_{\mathrm{nc}}/2^{i}\,:\, i=0,\dots,4\}} \mathbb{E}\big[(X - X')^2\big].
\label{eq:argmin}
\end{equation}

\subsection{Topology Dependency Resolution}
\label{sec:deps}

\begin{figure*}[t]
\centering
\includegraphics[width=0.95\textwidth]{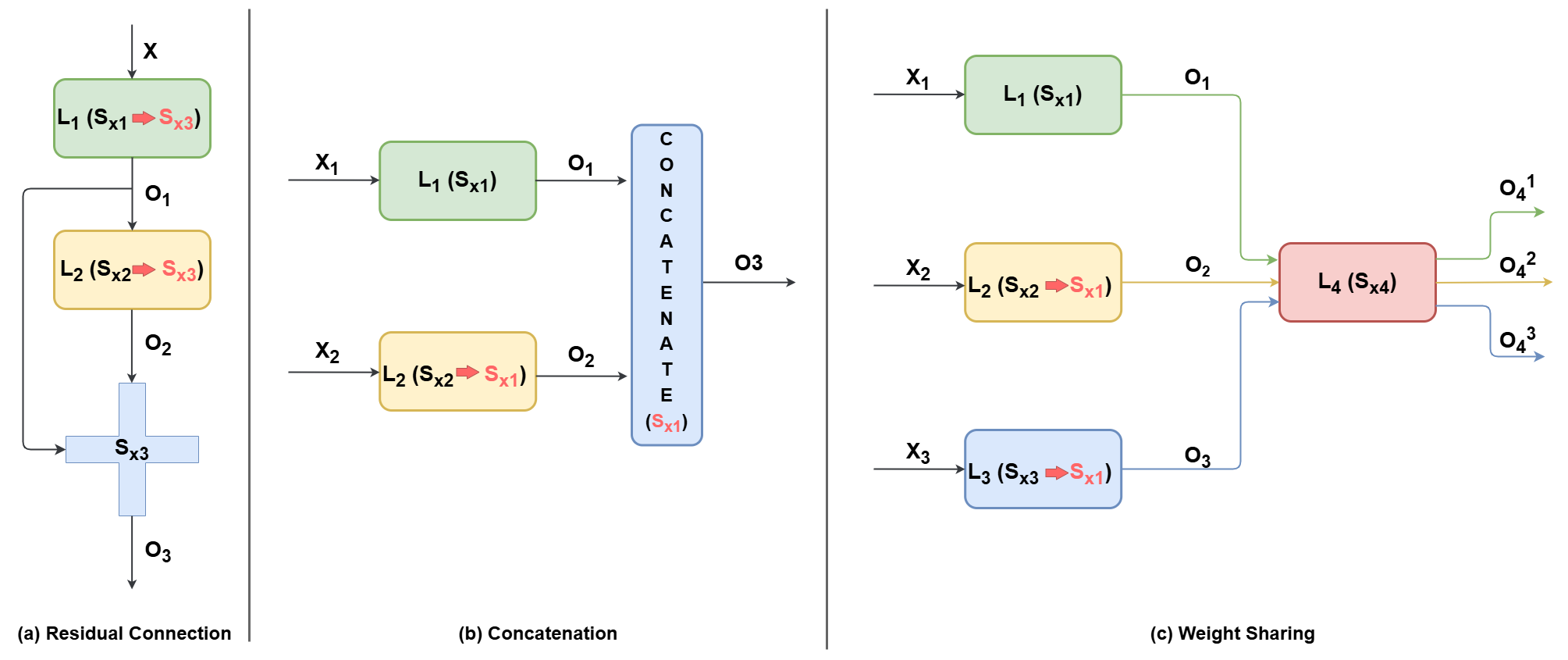}
\caption{Dependency resolution. (a) For a residual connection, both input layers adopt the scale of the element-wise addition. (b) For a concatenation, the inputs adopt the largest scale among them. (c) For weight sharing, all layers feeding the shared layer adopt the largest scale among them, so that one set of shared weights and one shift serve every invocation.}
\label{fig:deps}
\end{figure*}

Constraint C7 means that wherever two branches meet, their tensors must already share a scale, because the datapath cannot rescale one branch at runtime to match the other.
Independently proposed per-layer scales do not have this property, so they must be reconciled before propagation.
ShiftQAT resolves three cases (Figure~\ref{fig:deps}).

\textbf{Residual connections.}
For an element-wise addition with inputs from $L_1$ and $L_2$ (scales $S_{x_1}$, $S_{x_2}$) and output scale $S_{x_3}$, both inputs are reassigned $S_{x_3}$.
The two addends then share a grid, so the accelerator adds them directly and saturates the sum to the 8-bit range, with no runtime alignment.

\textbf{Concatenation.}
The output range of a concatenation is set by its widest input, so all inputs adopt the maximum of their proposed scales.
If $S_{x_1} > S_{x_2}$, then $S_{x_2}\leftarrow S_{x_1}$.

\textbf{Weight sharing.}
In FPN-style heads~\citep{DBLP:journals/corr/LinDGHHB16}, one layer $L_4$ may be invoked on inputs from $L_1$, $L_2$, and $L_3$ at different points in the network. A single set of weights and a single shift must then serve every invocation.
The scale $S_{x_4}$ is proposed from the union of $L_4$'s outputs across invocations, and the inputs $L_1$--$L_3$ are assigned the maximum of their scales.

Sharing scales at additions and concatenations follows the practice of TQT~\citep{TQT_2019}. ShiftQAT additionally resolves weight sharing and then folds all scales forward (Section~\ref{sec:prop}).
In each case the rule favors the larger scale, or the scale of the merge point.
This trades some resolution on the narrower branch for the guarantee that no branch clips.

\subsection{Static Scale Propagation}
\label{sec:prop}

\begin{figure}[t]
\centering
\includegraphics[width=\columnwidth]{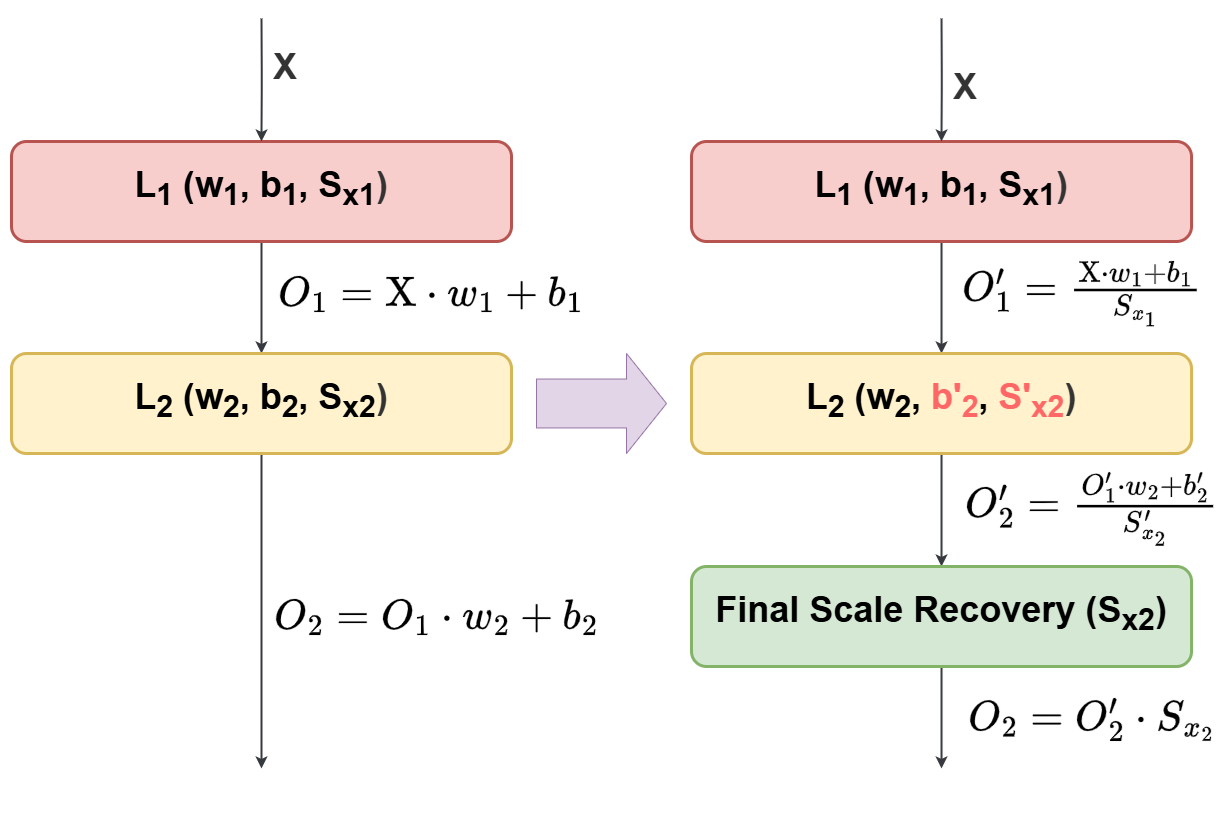}
\caption{Static scale propagation on a two-layer network. On the left, each layer's output carries its real-valued scale. On the right, each layer emits scale-normalized outputs, downstream biases and scales are adjusted offline (Eq.~\ref{eq:prop}), and a single recovery remains at the output. Activations are omitted from the diagram for clarity, but the derivation requires them to be positively homogeneous.}
\label{fig:prop}
\end{figure}

With a scale-consistent graph, each activation scale is folded into the next layer, so that no intermediate layer needs to recover real-valued scale (C6, C7).
Consider two layers with activation $f$, where $O_1 = f(Xw_1 + b_1)$ and $O_2 = f(O_1w_2 + b_2)$ (Figure~\ref{fig:prop}).
Layer~1 emits its output normalized by its own scale:
\begin{equation}
O_1' = \frac{f(Xw_1+b_1)}{S_{x_1}} = f\!\left(\frac{Xw_1+b_1}{S_{x_1}}\right),
\end{equation}
where the second equality holds because $f$ is positively homogeneous and $S_{x_1}>0$.
This is the only property of $f$ the method uses, and it is why Sigmoid and Tanh are excluded (Section~\ref{sec:limits}).
Layer~2 then consumes $O_1'$ directly, with its bias and scale adjusted offline:
\begin{equation}
b_2' = \frac{b_2}{S_{x_1}},\qquad S'_{x_2} = \frac{S_{x_2}}{S_{x_1}},
\label{eq:prop}
\end{equation}
\begin{equation}
O_2' = f\!\left(\frac{O_1'w_2 + b_2'}{S'_{x_2}}\right) = \frac{O_2}{S_{x_2}} .
\end{equation}
Since all scales are powers of two, $S'_{x_2}$ is also a power of two.
Applied layer by layer in topological order, this leaves a single recovery $O = O'\cdot S_{x}$ at the network output. The Arm core applies it when it reads out the results, and it can be omitted entirely for argmax classification. The last layer may instead emit unclipped 32-bit outputs, which avoids saturating the logits to 8 bits.

\subsection{Joint Weight/Bias Power-of-Two Scale}
\label{sec:wb}

Propagation (Eq.~\ref{eq:prop}) divides each bias by an upstream activation scale.
This changes the biases' dynamic range relative to the weights, and C4 requires the result to fit in 8 bits.
If weights and biases were scaled independently, the datapath would need a separate bias alignment, which violates C2 and C3.
ShiftQAT therefore quantizes each layer's weights and biases with a single shared power-of-two scale $S_w$, each at its own bit-width:
\begin{equation}
\Delta_w = \frac{S_w}{2^{n_w-1}},\qquad \Delta_b = \frac{S_w}{2^{n_b-1}},
\end{equation}
\begin{equation}
W' = \frac{\mathrm{clip}\!\left(\lfloor W/\Delta_w\rceil;\, -2^{n_w-1},\, 2^{n_w-1}-1\right)}{2^{n_w-1}},
\end{equation}
\begin{equation}
b' = \frac{\mathrm{clip}\!\left(\lfloor b/\Delta_b\rceil;\, -2^{n_b-1},\, 2^{n_b-1}-1\right)}{2^{n_b-1}},
\end{equation}
so that $W', b' \in [-1,1)$.
At every training iteration, weights and biases are pooled into one distribution, and outliers are rejected by taking the $q$-quantile of their absolute values as the range bound $\lambda$. We use $q=0.999$ for 8-bit layers and $q=0.9$ for 4-bit layers. With only 16 levels, clipping the largest 10\% of magnitudes buys more resolution for the bulk of the distribution than it costs in clipping error. The scale is then
\begin{equation}
S_w = 2^{s_w},\qquad s_w = \mathrm{clip}\big(\lceil \log_2 \lambda \rceil;\, -15,\, 15\big),
\label{eq:sw}
\end{equation}
where $\lambda$ is the quantile-based absolute-value bound and the clipping range follows the hardware's shift range (C2).
Because $S_w$ is recomputed as the parameters move, the network is trained on the grid it will be deployed on.
Recomputing $S_w$ instead of learning it avoids the extra machinery that learned power-of-two scales require to stay stable, such as threshold freezing and carefully bounded optimizer settings~\citep{TQT_2019}, and works with plain SGD.
When weights and biases compete for the same power-of-two range, training can shift magnitude between them rather than having one operand lose resolution at deployment.
Gradients pass through the rounding with the straight-through estimator~\citep{bengio2013_ste}.

\subsection{The PoT Scaler Is the Per-Layer Shift}
\label{sec:potscaler}

After quantization, a layer's accumulator $A = X'W' + b'$ carries the scale $S_w$, while its output must carry the propagated scale $S'_x$.
The required rescaling is therefore
\begin{equation}
S_{\mathrm{PoT}} = \frac{S_w}{S'_x} = 2^{k},\qquad k = \mathrm{clip}\big(s_w - s'_x;\, -15,\, 15\big),
\label{eq:pot}
\end{equation}
where $s'_x = \log_2 S'_x$.
Multiplying $A$ by $2^{k}$ is a single arithmetic shift, to the left for $k>0$ and to the right for $k<0$.
The hardware applies it to the 32-bit intermediate result and accepts total shifts in $[-15,15]$, so the flow clips $k$ to that range during training.
The activation is then applied, and the result is quantized with step $2^{-(n_x-1)}$ and saturated to $[-1,\,1-2^{-(n_x-1)}]$.
On MAX78002, this value is exactly the accelerator's programmable per-layer output shift. The weight-quantization shift and the propagated activation-scale shift are merged into each layer's single shift value.
The PoT scaler is therefore not an extra operation that ShiftQAT adds.
It is the training-time computation of the value the hardware register will be programmed with, and QAT keeps that value consistent with the evolving weights at every iteration.

\section{Evaluation}
\label{sec:results}

\subsection{Setup}
We quantize MobileNetV2 and ResNet-18 starting from the official PyTorch FP32 checkpoints~\citep{Ansel_PyTorch_2_Faster_2024}, trained on ImageNet at 224$\times$224~\citep{imagenet_cvpr09}.
We evaluate two settings, W8A8 and W4A8. For a fair comparison with published results, all methods in Tables~\ref{tab:w8a8} and~\ref{tab:w4a8}, including ShiftQAT, use 32-bit biases ($n_b=32$), as is standard in the quantization literature. These tables therefore isolate the effect of the weight and activation constraints (C1--C3, C5--C7). The deployed networks in Section~\ref{sec:hw} use 8-bit biases ($n_b=8$), as the hardware requires (C4).
QAT runs for the last 50 of 200 epochs for MobileNetV2 and for the last 30 of 120 epochs for ResNet-18.
Appendix~\ref{app:exp} gives the full hyperparameters.
All ShiftQAT results come from a single run per configuration.

Baseline numbers are taken from the cited papers.
Because the baselines use different FP32 references (e.g., 71.878\% for our MobileNetV2 checkpoint versus 72.62\% in EPTQ's), we compare accuracy drops ($\Delta$) rather than absolute accuracy.
We treat differences below about 1~pp as comparable, given single runs and cross-paper references.
We also mark each method's regime, post-training (PTQ) or quantization-aware (QAT), because PTQ methods use far less compute than a QAT schedule.
No baseline can be deployed on MAX78002 without violating part of the contract, so all comparisons are accuracy-only.

\subsection{Accuracy Under the Contract's Scaling Constraints}
\label{sec:acc}

\begin{table*}[t]
\centering
\small
\setlength{\tabcolsep}{4.2pt}
\renewcommand{\arraystretch}{1.12}
\caption{W8A8 ImageNet top-1 accuracy. Baseline MobileNetV2 results are as reported by HPTQ~\citep{habi_2021_hptq}, and baseline ResNet-18 results are as reported by LSQ~\citep{esser_2020_lsq}. A ``$+$'' marks a satisfied property and a ``$-$'' an unsatisfied one. The Static prop. column marks the static scale propagation of Section~\ref{sec:prop}, which is this paper's mechanism and so is satisfied only by ShiftQAT. The Regime column gives post-training (PTQ) or quantization-aware training (QAT). ShiftQAT is the only row satisfying every property column. All methods use 32-bit biases.}
\label{tab:w8a8}
\begin{tabular}{llcccccccccc}
\toprule
\textbf{Model} & \textbf{Method} & \textbf{Regime} & \textbf{FP (\%)} & \textbf{Q (\%)} & \textbf{$\Delta$ (pp)} & \textbf{Unif.} & \textbf{Sym.} & \textbf{PoT} & \textbf{Per-ten.} & \textbf{Static prop.} \\
\midrule
MobileNetV2 & ShiftQAT & QAT & 71.878 & 70.852 & 1.026 & $+$ & $+$ & $+$ & $+$ & $+$ \\
 & TQT~\citep{TQT_2019} & QAT & 71.7 & 71.8 & $-0.1$ & $+$ & $+$ & $+$ & $+$ & $-$ \\
 & QAT~\citep{jacob_2018_CVPR} & QAT & 71.9 & 70.9 & 1.0 & $+$ & $-$ & $-$ & $+$ & $-$ \\
 & HPTQ~\citep{habi_2021_hptq} & PTQ & 71.812 & 71.46 & 0.352 & $+$ & $+$ & $+$ & $-$ & $-$ \\
\midrule
ResNet-18 & ShiftQAT & QAT & 69.758 & 69.2 & 0.558 & $+$ & $+$ & $+$ & $+$ & $+$ \\
 & LSQ~\citep{esser_2020_lsq} & QAT & 70.5 & 71.1 & $-0.6$ & $+$ & $+$ & $-$ & $+$ & $-$ \\
 & FAQ~\citep{mckinstry_2019_faq} & QAT & 70.5 & 70.0 & 0.5 & $+$ & $+$ & $+$ & $+$ & $-$ \\
\bottomrule
\end{tabular}
\end{table*}

\begin{table*}[t]
\centering
\small
\setlength{\tabcolsep}{4.2pt}
\renewcommand{\arraystretch}{1.12}
\caption{W4A8 ImageNet top-1 accuracy. PTQ baseline results are as reported by EPTQ~\citep{gordon_2024_eptq}. QAT baseline results are the per-tensor results of \citet{nagel_2021_whitepaper} (average of three runs). Columns as in Table~\ref{tab:w8a8}. All methods use 32-bit biases.}
\label{tab:w4a8}
\begin{tabular}{llcccccccccc}
\toprule
\textbf{Model} & \textbf{Method} & \textbf{Regime} & \textbf{FP (\%)} & \textbf{Q (\%)} & \textbf{$\Delta$ (pp)} & \textbf{Unif.} & \textbf{Sym.} & \textbf{PoT} & \textbf{Per-ten.} & \textbf{Static prop.} \\
\midrule
MobileNetV2 & ShiftQAT & QAT & 71.878 & 69.014 & 2.864 & $+$ & $+$ & $+$ & $+$ & $+$ \\
 & QAT~\citep{nagel_2021_whitepaper} & QAT & 71.72 & 70.17 & 1.55 & $+$ & $-$ & $-$ & $+$ & $-$ \\
 & AdaRound~\citep{markus2020adaround} & PTQ & 72.62 & 69.25 & 3.37 & $+$ & $-$ & $-$ & $+$ & $-$ \\
 & BRECQ~\citep{li2021brecq} & PTQ & 72.62 & 71.42 & 1.2 & $+$ & $-$ & $-$ & $-$ & $-$ \\
 & EPTQ~\citep{gordon_2024_eptq} & PTQ & 72.62 & 72.06 & 0.56 & $+$ & $+$ & $-$ & $-$ & $-$ \\
\midrule
ResNet-18 & ShiftQAT & QAT & 69.758 & 67.854 & 1.904 & $+$ & $+$ & $+$ & $+$ & $+$ \\
 & QAT~\citep{nagel_2021_whitepaper} & QAT & 69.68 & 69.76 & $-0.08$ & $+$ & $-$ & $-$ & $+$ & $-$ \\
 & AdaRound~\citep{markus2020adaround} & PTQ & 71.01 & 68.55 & 2.46 & $+$ & $-$ & $-$ & $+$ & $-$ \\
 & BRECQ~\citep{li2021brecq} & PTQ & 71.01 & 70.58 & 0.43 & $+$ & $-$ & $-$ & $-$ & $-$ \\
 & EPTQ~\citep{gordon_2024_eptq} & PTQ & 71.01 & 70.70 & 0.31 & $+$ & $+$ & $-$ & $-$ & $-$ \\
\bottomrule
\end{tabular}
\end{table*}

\textbf{W8A8 (Table~\ref{tab:w8a8}).}
ShiftQAT loses 1.026~pp on MobileNetV2 and 0.558~pp on ResNet-18.
These drops match those of integer-only QAT~\citep{jacob_2018_CVPR} on MobileNetV2 (1.0~pp), which uses asymmetric, non-power-of-two scaling, and of FAQ~\citep{mckinstry_2019_faq} on ResNet-18 (0.5~pp).
At 8 bits, then, meeting all of the contract's scaling constraints costs no measurable accuracy relative to comparable QAT methods that meet only part of them.
HPTQ and LSQ report smaller drops, but HPTQ relies on per-channel scales and LSQ on non-power-of-two step sizes, neither of which the datapath can execute.
TQT, which is also power-of-two and per-tensor, reports a 0.1~pp gain at W8A8.
However, its results table leaves the W4A8 (INT4/INT8) entry blank for both MobileNetV1 and MobileNetV2, the only networks in that table without a reported INT4 result~\citep{TQT_2019}.
ShiftQAT, by contrast, delivers W4A8 MobileNetV2 within 2.9~pp of FP32 under stricter constraints.

\textbf{W4A8 (Table~\ref{tab:w4a8}).}
With 4-bit weights, ShiftQAT loses 2.864~pp on MobileNetV2 and 1.904~pp on ResNet-18.
This is less than AdaRound (3.37 and 2.46~pp), the per-tensor PTQ method in the table.
Per-tensor QAT with learned scales~\citep{nagel_2021_whitepaper} loses less, 1.55 and $-0.08$~pp.
It relaxes parts of the contract that ShiftQAT must keep.
It uses asymmetric activations, which give post-ReLU tensors all 8 bits where C5 leaves 7, and it learns non-power-of-two scales by backpropagation, whereas ShiftQAT keeps power-of-two activation scales fixed after calibration.
The remaining gap is therefore the observed price of the full scaling contract at 4 bits, and separating the contribution of each factor is left to future work (Section~\ref{sec:limits}).
BRECQ and EPTQ reach smaller drops with per-channel, non-power-of-two scales, which violate C2 and C3.
ShiftQAT is the only method in the table whose scaling scheme the datapath can execute. With biases constrained to 8 bits, the same recipe produces the deployed networks of Section~\ref{sec:hw}.
Differences across rows also reflect differing FP32 references and training regimes.

\subsection{End-to-End Deployment on MAX78002}
\label{sec:hw}

MAX78002 supports 1/2/4/8-bit weights with 8-bit activations.
We deploy MobileNetV2 and ResNet-18 trained with the same ShiftQAT recipe, with mixed 4/8-bit weights and 8-bit activations. Unlike in Tables~\ref{tab:w8a8} and~\ref{tab:w4a8}, the biases are quantized to 8 bits, as the hardware requires (C4), so these networks satisfy the full contract.
The trained checkpoints are compiled with ai8x-synthesis, the toolchain's open-source synthesis tool, with no additional calibration.
Because full-size ImageNet models exceed the accelerator's on-chip weight memory, the deployed variants are reduced.
MobileNetV2 uses width multiplier $m=0.75$ and ResNet-18 uses $m=0.5$, both with 128$\times$128 inputs. Layers wider than 1024 channels are capped at 1024, and each stride-2 convolution is replaced by 2$\times$2 max-pooling followed by a stride-1 convolution (Appendix~\ref{app:hwimpl}).
The mixed-precision assignment keeps the input and output layers at 8-bit weights and concentrates the 4-bit layers in the interior of the network.
This follows a long-standing convention in low-bit quantization, where the first and last layers are kept at higher precision because they are the most sensitive to quantization error~\citep{rastegari_2016_xnor, zhou2018dorefanettraininglowbitwidth, choi_2018_pact, esser_2020_lsq}.
TQT's authors likewise attribute part of their INT4 accuracy shortfall to not keeping the first and last layers at higher precision~\citep{TQT_2019}. Hessian-based analyses also confirm that sensitivity to quantization varies strongly across layers~\citep{dong_2019_hawq}.
The input layer sees raw pixels through only three input channels and holds few parameters, so 8 bits there cost little memory, while the output layer directly produces the logits.
Latency and energy are measured with the MAX78002 evaluation kit's on-board power monitor in its windowed energy-accumulation (CNN power) mode.
Firmware instrumentation, generated by the synthesis tool, marks idle and active windows with trigger signals.
The monitor accumulates power on the VCOREA rail, which supplies the CNN accelerator as well as the MCU logic, and times the active window.
It subtracts the idle power, so that $E=(A-I)\,T$, where $A$ and $I$ are the active and idle power and $T$ is the active period. Each measured operation is averaged over 100 repetitions.
Weight loading is measured as a separate stage and is not included in Table~\ref{tab:hw}.
The CNN accelerator runs at 200~MHz. Table~\ref{tab:hw} reports inference only, obtained by subtracting the input-loading stage from the input-loading-plus-inference stage.

\begin{table*}[t]
\centering
\small
\setlength{\tabcolsep}{4pt}
\renewcommand{\arraystretch}{1.15}
\caption{MobileNetV2 and ResNet-18 deployed end-to-end on MAX78002 with the ShiftQAT flow (CNN clock 200~MHz). Accuracy is ImageNet top-1. Time and energy are per inference, excluding weight and input loading, measured on the VCOREA rail with idle power subtracted.}
\label{tab:hw}
\begin{tabular}{lcccccccc}
\toprule
\multirow{2}{*}{\textbf{Model}} & \multirow{2}{*}{\textbf{Input}} & \multirow{2}{*}{\textbf{Acc. (\%)}} & \multicolumn{2}{c}{\textbf{Weights}} & \multicolumn{2}{c}{\textbf{Biases}} & \multicolumn{2}{c}{\textbf{Inference}} \\
\cmidrule(lr){4-5}\cmidrule(lr){6-7}\cmidrule(lr){8-9}
 & & & \textbf{Num.\,(M)} & \textbf{Size\,(MB)} & \textbf{Num.\,(k)} & \textbf{Size\,(kB)} & \textbf{Time\,(ms)} & \textbf{Energy\,($\mu$J)} \\
\midrule
MobileNetV2 ($m{=}0.75$) & 128$\times$128 & 62.812 & 2.29 & 2.03 & 6.6 & 6.6 & 14.6 & 5137 \\
ResNet-18 ($m{=}0.5$) & 128$\times$128 & 53.230 & 3.06 & 2.05 & 3.4 & 3.4 & 8.4 & 2768 \\
\bottomrule
\end{tabular}
\end{table*}

Table~\ref{tab:hw} reports the results.
The deployed MobileNetV2 runs in 14.6~ms (about 68 inferences per second) at 5.1~mJ per inference, and ResNet-18 runs in 8.4~ms (about 119 inferences per second) at 2.8~mJ.
Every layer executes on the accelerator in integer arithmetic, and only the final output rescale runs on the Arm core. These are ImageNet-scale CNNs running on a microcontroller-class device at millisecond latency and millijoule energy.
The memory columns show what drove the model design.
Both variants store about 2~MB of weights, most of MAX78002's 2,340~KiB weight memory, whose effective capacity is further reduced for layers with more than 64 channels.
On average they store roughly 7 bits per weight for MobileNetV2 and roughly 5.5 bits for ResNet-18, so the 4-bit option was used far more heavily for ResNet-18.
All biases occupy exactly one byte each, as C4 requires.

\textbf{Reduced variants.}
To fit in on-chip weight memory, the deployed networks are narrower than the full-size models, take 128$\times$128 inputs, and use mixed 4/8-bit weights. They also use 8-bit rather than 32-bit biases.
Their accuracy is therefore not directly comparable to the full-size results in Section~\ref{sec:acc}.
Table~\ref{tab:hw} demonstrates what the flow enables on silicon, not a controlled accuracy comparison.

\section{Lessons from Building and Deploying the Flow}
\label{sec:lessons}

\textbf{L1. The contract, not the bit-width, is the specification.}
``W4A8'' underdetermines a deployable network.
Two W4A8 networks can differ in symmetry, scale granularity, bias width, and where rescaling happens, and only one of them may compile.
Writing the contract down explicitly (Table~\ref{tab:contract}) before designing the training flow is what made each mechanism in Section~\ref{sec:method} necessary rather than optional.

\textbf{L2. Constraints interact, so satisfy them jointly.}
Each constraint in isolation has a known solution.
The difficulty comes from their interaction.
Folding activation scales forward (C6, C7) moves them into biases, which must then fit in 8 bits (C4) on the same power-of-two grid as the weights (C2).
The joint, per-iteration weight/bias scale exists only because of this interaction.

\textbf{L3. Topology, not arithmetic, is where static scaling breaks.}
The propagation identity itself (Eq.~\ref{eq:prop}) is a few lines of algebra.
Making it valid on real networks required resolving every point where branches merge or weights are reused (Section~\ref{sec:deps}).
Any flow targeting a datapath without dynamic cross-branch rescaling must solve this before it can compile ResNet- or FPN-style graphs.

\textbf{L4. On-chip weight memory shapes the deployed model.}
The deployed models were sized to fill the weight memory (Table~\ref{tab:hw}).
Width, input resolution, the 1024-channel limit, and the 4/8-bit mix were all chosen so that the network fits.
Wider or higher-resolution deployments come from more weight memory or lower average bits per weight, and a QAT flow's leverage lies in making the lower-bit option accurate enough to use. ShiftQAT's W4A8 results are what make 4-bit interior layers a practical way to fit larger networks.

\textbf{L5. Make training emit the hardware configuration.}
The PoT scaler (Section~\ref{sec:potscaler}) computes, at every iteration, the exact value that each layer's output-shift register will hold.
A trained checkpoint therefore maps deterministically to integer weights, 8-bit biases, and shift values.
No post-training calibration step is needed, so the hardware never runs a configuration that training did not see.
In practice, we found this property more valuable than any single accuracy trick, because it makes the path from training to silicon predictable.

\section{Discussion and Scope}
\label{sec:limits}

\textbf{Beyond MAX7800x.}
The constraints in Table~\ref{tab:contract} are typical of shift-based fixed-point datapaths, and the mechanisms of Section~\ref{sec:method} do not depend on MAX7800x specifics beyond them.
On accelerators that lack a free per-layer shifter, static propagation would additionally remove per-layer rescaling from the datapath.
On MAX78002, whose output shift is free, its value lies in making the network executable at all.

\textbf{Closing the 4-bit gap.}
Learned power-of-two thresholds, as in TQT~\citep{TQT_2019}, remain compatible with the contract, because the learned scales are still powers of two. Adopting them in ShiftQAT is a direct way to narrow the W4A8 gap to QAT methods that relax the contract.

\textbf{Activation functions.}
Propagation requires every activation between the input and the output to be positively homogeneous, such as ReLU, Leaky ReLU~\citep{maas2013rectifier}, or the modulus~\citep{perez_2023_modulus_activation}.
These activations are standard in edge CNNs.
Networks with Sigmoid, Tanh, GELU, or SiLU would need scale recovery at those layers, which a static datapath cannot provide.

\section{Conclusion}

Fixed-function edge accelerators define a numerical contract, and a QAT flow for them must be designed to that contract rather than to a bit-width.
ShiftQAT meets the full contract of the MAX7800x family. It resolves topology dependencies so that activation scales propagate statically, and it trains weights and biases on a shared power-of-two grid that fits an 8-bit bias budget.
Under the contract's scaling constraints, it matches the W8A8 accuracy drop of less constrained QAT methods and stays within 2.9~pp of FP32 at W4A8.
With 8-bit biases, ShiftQAT-trained networks run end-to-end on MAX78002 at millisecond latency and millijoule energy per inference.
The flow is open source and is the default training path of the MAX7800x toolchain, and we expect its constraint-to-mechanism design to carry over to other fixed-function accelerators.

\bibliography{main}

@incollection{gholami2022survey,
  title={A survey of quantization methods for efficient neural network inference},
  author={Gholami, Amir and Kim, Sehoon and Dong, Zhen and Yao, Zhewei and Mahoney, Michael W and Keutzer, Kurt},
  booktitle={Low-Power Computer Vision},
  pages={291--326},
  year={2022},
  publisher={Chapman and Hall/CRC}
}

@inproceedings{hubara_courbariauc_nips2016,
 author = {Hubara, Itay and Courbariaux, Matthieu and Soudry, Daniel and El-Yaniv, Ran and Bengio, Yoshua},
 booktitle = {Advances in Neural Information Processing Systems},
 editor = {D. Lee and M. Sugiyama and U. Luxburg and I. Guyon and R. Garnett},
 pages = {},
 publisher = {Curran Associates, Inc.},
 title = {Binarized Neural Networks},
 url = {https://proceedings.neurips.cc/paper_files/paper/2016/file/d8330f857a17c53d217014ee776bfd50-Paper.pdf},
 volume = {29},
 year = {2016}
}

@misc{bengio2013_ste,
      title={Estimating or Propagating Gradients Through Stochastic Neurons for Conditional Computation}, 
      author={Yoshua Bengio and Nicholas Léonard and Aaron Courville},
      year={2013},
      eprint={1308.3432},
      archivePrefix={arXiv},
      primaryClass={cs.LG},
      url={https://arxiv.org/abs/1308.3432}, 
}

@misc{zhou2018dorefanettraininglowbitwidth,
      title={DoReFa-Net: Training Low Bitwidth Convolutional Neural Networks with Low Bitwidth Gradients}, 
      author={Shuchang Zhou and Yuxin Wu and Zekun Ni and Xinyu Zhou and He Wen and Yuheng Zou},
      year={2018},
      eprint={1606.06160},
      archivePrefix={arXiv},
      primaryClass={cs.NE},
      url={https://arxiv.org/abs/1606.06160}, 
}

@InProceedings{zhang_2018_eccv_lqnets,
author = {Zhang, Dongqing and Yang, Jiaolong and Ye, Dongqiangzi and Hua, Gang},
title = {LQ-Nets: Learned Quantization for Highly Accurate and Compact Deep Neural Networks},
booktitle = {Proceedings of the European Conference on Computer Vision (ECCV)},
month = {September},
year = {2018}
}

@InProceedings{jung_2019_cvpr,
author = {Jung, Sangil and Son, Changyong and Lee, Seohyung and Son, Jinwoo and Han, Jae-Joon and Kwak, Youngjun and Hwang, Sung Ju and Choi, Changkyu},
title = {Learning to Quantize Deep Networks by Optimizing Quantization Intervals With Task Loss},
booktitle = {Proceedings of the IEEE/CVF Conference on Computer Vision and Pattern Recognition (CVPR)},
month = {June},
year = {2019}
}

@inproceedings{yang_2021_iclr_bsq,
title={{\{}BSQ{\}}: Exploring Bit-Level Sparsity for Mixed-Precision Neural Network Quantization},
author={Huanrui Yang and Lin Duan and Yiran Chen and Hai Li},
booktitle={International Conference on Learning Representations},
year={2021},
url={https://openreview.net/forum?id=TiXl51SCNw8}
}

@misc{habi_2021_hptq,
      title={HPTQ: Hardware-Friendly Post Training Quantization}, 
      author={Hai Victor Habi and Reuven Peretz and Elad Cohen and Lior Dikstein and Oranit Dror and Idit Diamant and Roy H. Jennings and Arnon Netzer},
      year={2021},
      eprint={2109.09113},
      archivePrefix={arXiv},
      primaryClass={cs.CV},
      url={https://arxiv.org/abs/2109.09113}, 
}

@inproceedings{migacz_2017,
  title={8-bit inference with tensorrt},
  author={Migacz, Szymon},
  booktitle={GPU technology conference},
  volume={2},
  issue ={4},
  pages={5},
  year={2017}
}

@InProceedings{jacob_2018_CVPR,
author = {Jacob, Benoit and Kligys, Skirmantas and Chen, Bo and Zhu, Menglong and Tang, Matthew and Howard, Andrew and Adam, Hartwig and Kalenichenko, Dmitry},
title = {Quantization and Training of Neural Networks for Efficient Integer-Arithmetic-Only Inference},
booktitle = {Proceedings of the IEEE Conference on Computer Vision and Pattern Recognition (CVPR)},
month = {June},
year = {2018}
}

@misc{mckinstry_2019_faq,
      title={Discovering Low-Precision Networks Close to Full-Precision Networks for Efficient Embedded Inference}, 
      author={Jeffrey L. McKinstry and Steven K. Esser and Rathinakumar Appuswamy and Deepika Bablani and John V. Arthur and Izzet B. Yildiz and Dharmendra S. Modha},
      year={2019},
      eprint={1809.04191},
      archivePrefix={arXiv},
      primaryClass={cs.CV},
      url={https://arxiv.org/abs/1809.04191}, 
}

@misc{choi_2018_pact,
      title={PACT: Parameterized Clipping Activation for Quantized Neural Networks}, 
      author={Jungwook Choi and Zhuo Wang and Swagath Venkataramani and Pierce I-Jen Chuang and Vijayalakshmi Srinivasan and Kailash Gopalakrishnan},
      year={2018},
      eprint={1805.06085},
      archivePrefix={arXiv},
      primaryClass={cs.CV},
      url={https://arxiv.org/abs/1805.06085}, 
}

@inproceedings{esser_2020_lsq,
    title={LEARNED STEP SIZE QUANTIZATION},
    author={Steven K. Esser and Jeffrey L. McKinstry and Deepika Bablani and Rathinakumar Appuswamy and Dharmendra S. Modha},
    booktitle={International Conference on Learning Representations},
    year={2020},
    url={https://openreview.net/forum?id=rkgO66VKDS}
}

@inproceedings{
    uhlich_2020_clr,
    title={Mixed Precision DNNs: All you need is a good parametrization},
    author={Stefan Uhlich and Lukas Mauch and Fabien Cardinaux and Kazuki Yoshiyama and Javier Alonso Garcia and Stephen Tiedemann and Thomas Kemp and Akira Nakamura},
    booktitle={International Conference on Learning Representations},
    year={2020},
    url={https://openreview.net/forum?id=Hyx0slrFvH}
}

@misc{gordon_2024_eptq,
      title={EPTQ: Enhanced Post-Training Quantization via Hessian-guided Network-wise Optimization}, 
      author={Ofir Gordon and Elad Cohen and Hai Victor Habi and Arnon Netzer},
      year={2024},
      eprint={2309.11531},
      archivePrefix={arXiv},
      primaryClass={cs.CV},
      url={https://arxiv.org/abs/2309.11531}, 
}

@misc{max78002,
  author = {{Analog Devices Inc.}},
  title = {{MAX78002} Datasheet and Product Info},
  year = {2023},
  howpublished = {\url{https://www.analog.com/en/products/max78002.html}},
  note = {Accessed: 2026-06-17},
}

@INPROCEEDINGS{giordano_2022_aicas,
  author={Giordano, Marco and Piccinelli, Luigi and Magno, Michele},
  booktitle={2022 IEEE 4th International Conference on Artificial Intelligence Circuits and Systems (AICAS)}, 
  title={Survey and Comparison of Milliwatts Micro controllers for Tiny Machine Learning at the Edge}, 
  year={2022},
  volume={},
  number={},
  pages={94-97},
  doi={10.1109/AICAS54282.2022.9870017}}

@INPROCEEDINGS{moosmann_2023_tinyssimoyolo,
  author={Moosmann, Julian and Giordano, Marco and Vogt, Christian and Magno, Michele},
  booktitle={2023 IEEE 5th International Conference on Artificial Intelligence Circuits and Systems (AICAS)}, 
  title={TinyissimoYOLO: A Quantized, Low-Memory Footprint, TinyML Object Detection Network for Low Power Microcontrollers}, 
  year={2023},
  volume={},
  number={},
  pages={1-5},
  doi={10.1109/AICAS57966.2023.10168657}}

@inproceedings{huang_2024_energycharacterization,
author = {Huang, Yushan and Gong, Taesik and Jang, SiYoung and Kawsar, Fahim and Min, Chulhong},
title = {Energy Characterization of Tiny AI Accelerator-Equipped Microcontrollers},
year = {2024},
isbn = {9798400713002},
publisher = {Association for Computing Machinery},
address = {New York, NY, USA},
url = {https://doi.org/10.1145/3698388.3699628},
doi = {10.1145/3698388.3699628},
booktitle = {Proceedings of the 2nd International Workshop on Human-Centered Sensing, Networking, and Multi-Device Systems},
pages = {1–6},
numpages = {6},
location = {Hangzhou, China},
series = {HumanSys '24}
}

@inproceedings{TQT_2019,
  author    = {Jain, Sambhav R. and Gural, Albert and Wu, Michael and Dick, Chris},
  title     = {Trained Quantization Thresholds for Accurate and Efficient Fixed-Point Inference of Deep Neural Networks},
  booktitle = {Proceedings of Machine Learning and Systems (MLSys)},
  volume    = {2},
  pages     = {112--128},
  year      = {2020},
}

@inproceedings{BED_acm_2022,
author = {Wang, Guanchu and Bhat, Zaid Pervaiz and Jiang, Zhimeng and Chen, Yi-Wei and Zha, Daochen and Reyes, Alfredo Costilla and Niktash, Afshin and Ulkar, Gorkem and Okman, Erman and Cai, Xuanting and Hu, Xia},
title = {BED: A Real-Time Object Detection System for Edge Devices},
year = {2022},
isbn = {9781450392365},
publisher = {Association for Computing Machinery},
address = {New York, NY, USA},
url = {https://doi.org/10.1145/3511808.3557168},
doi = {10.1145/3511808.3557168},
booktitle = {Proceedings of the 31st ACM International Conference on Information \& Knowledge Management},
pages = {4994–4998},
numpages = {5},
location = {Atlanta, GA, USA},
series = {CIKM '22}
}

@InProceedings{dabbous_2025_tinyml_max78000,
author="Dabbous, Ali
and Lazzaroni, Luca
and Bellotti, Francesco
and Presentaci{\'o}n, Sara Mu{\~{n}}oz
and Pighetti, Alessandro
and Berta, Riccardo",
editor="Valle, Maurizio
and Gastaldo, Paolo
and Limiti, Ernesto",
title="TinyML Acceleration with MAX78000",
booktitle="Proceedings of SIE 2024",
year="2025",
publisher="Springer Nature Switzerland",
address="Cham",
pages="468--474",
isbn="978-3-031-71518-1"
}

@article{ioffe2015batch,
  title={Batch normalization: Accelerating deep network training by reducing internal covariate shift},
  author={Ioffe, Sergey},
  journal={arXiv preprint arXiv:1502.03167},
  year={2015}
}

@inproceedings{maas2013rectifier,
  title={Rectifier nonlinearities improve neural network acoustic models},
  author={Maas, Andrew L and Hannun, Awni Y and Ng, Andrew Y},
  booktitle={Proc. icml},
  volume={30},
  issue ={1},
  pages={3},
  year={2013},
  organization={Atlanta, GA}
}

@article{li2021brecq,
  title={Brecq: Pushing the limit of post-training quantization by block reconstruction},
  author={Li, Yuhang and Gong, Ruihao and Tan, Xu and Yang, Yang and Hu, Peng and Zhang, Qi and Yu, Fengwei and Wang, Wei and Gu, Shi},
  journal={arXiv preprint arXiv:2102.05426},
  year={2021}
}

@article{markus2020adaround,
  author       = {Markus Nagel and
                  Rana Ali Amjad and
                  Mart van Baalen and
                  Christos Louizos and
                  Tijmen Blankevoort},
  title        = {Up or Down? Adaptive Rounding for Post-Training Quantization},
  journal      = {CoRR},
  volume       = {abs/2004.10568},
  year         = {2020},
  url          = {https://arxiv.org/abs/2004.10568},
  eprinttype    = {arXiv},
  eprint       = {2004.10568},
  bibsource    = {dblp computer science bibliography, https://dblp.org}
}

@inproceedings{Ansel_PyTorch_2_Faster_2024,
author = {Ansel, Jason and Yang, Edward and He, Horace and Gimelshein, Natalia and Jain, Animesh and Voznesensky, Michael and Bao, Bin and Bell, Peter and Berard, David and Burovski, Evgeni and Chauhan, Geeta and Chourdia, Anjali and Constable, Will and Desmaison, Alban and DeVito, Zachary and Ellison, Elias and Feng, Will and Gong, Jiong and Gschwind, Michael and Hirsh, Brian and Huang, Sherlock and Kalambarkar, Kshiteej and Kirsch, Laurent and Lazos, Michael and Lezcano, Mario and Liang, Yanbo and Liang, Jason and Lu, Yinghai and Luk, CK and Maher, Bert and Pan, Yunjie and Puhrsch, Christian and Reso, Matthias and Saroufim, Mark and Siraichi, Marcos Yukio and Suk, Helen and Suo, Michael and Tillet, Phil and Wang, Eikan and Wang, Xiaodong and Wen, William and Zhang, Shunting and Zhao, Xu and Zhou, Keren and Zou, Richard and Mathews, Ajit and Chanan, Gregory and Wu, Peng and Chintala, Soumith},
booktitle = {29th ACM International Conference on Architectural Support for Programming Languages and Operating Systems, Volume 2 (ASPLOS '24)},
doi = {10.1145/3620665.3640366},
month = {apr},
publisher = {ACM},
title = {{PyTorch 2: Faster Machine Learning Through Dynamic Python Bytecode Transformation and Graph Compilation}},
url = {https://pytorch.org/assets/pytorch2-2.pdf},
year = {2024}
}

@article{perez_2023_modulus_activation,
title = {Empirical study of the modulus as activation function in computer vision applications},
journal = {Engineering Applications of Artificial Intelligence},
volume = {120},
pages = {105863},
year = {2023},
issn = {0952-1976},
doi = {https://doi.org/10.1016/j.engappai.2023.105863},
url = {https://www.sciencedirect.com/science/article/pii/S0952197623000477},
author = {Iván Vallés-Pérez and Emilio Soria-Olivas and Marcelino Martínez-Sober and Antonio J. Serrano-López and Joan Vila-Francés and Juan Gómez-Sanchís},
}

@inproceedings{imagenet_cvpr09,
        AUTHOR = {Deng, J. and Dong, W. and Socher, R. and Li, L.-J. and Li, K. and Fei-Fei, L.},
        TITLE = {{ImageNet: A Large-Scale Hierarchical Image Database}},
        BOOKTITLE = {CVPR09},
        YEAR = {2009},
        BIBSOURCE = "http://www.image-net.org/papers/imagenet_cvpr09.bib"}

@article{DBLP:journals/corr/LinDGHHB16,
  author       = {Tsung{-}Yi Lin and
                  Piotr Doll{\'{a}}r and
                  Ross B. Girshick and
                  Kaiming He and
                  Bharath Hariharan and
                  Serge J. Belongie},
  title        = {Feature Pyramid Networks for Object Detection},
  journal      = {CoRR},
  volume       = {abs/1612.03144},
  year         = {2016},
  url          = {http://arxiv.org/abs/1612.03144},
  eprinttype    = {arXiv},
  eprint       = {1612.03144},
  bibsource    = {dblp computer science bibliography, https://dblp.org}
}

@inproceedings{nagel_2019_dfq,
  author    = {Nagel, Markus and van Baalen, Mart and Blankevoort, Tijmen and Welling, Max},
  title     = {Data-Free Quantization Through Weight Equalization and Bias Correction},
  booktitle = {Proceedings of the IEEE/CVF International Conference on Computer Vision (ICCV)},
  pages     = {1325--1334},
  year      = {2019},
}

@inproceedings{yao_2021_hawqv3,
  author    = {Yao, Zhewei and Dong, Zhen and Zheng, Zhangcheng and Gholami, Amir and Yu, Jiali and Tan, Eric and Wang, Leyuan and Huang, Qijing and Wang, Yida and Mahoney, Michael W. and Keutzer, Kurt},
  title     = {{HAWQ-V3}: Dyadic Neural Network Quantization},
  booktitle = {Proceedings of the 38th International Conference on Machine Learning (ICML)},
  series    = {PMLR},
  volume    = {139},
  pages     = {11875--11886},
  year      = {2021},
}

@inproceedings{li_2020_apot,
  author    = {Li, Yuhang and Dong, Xin and Wang, Wei},
  title     = {Additive Powers-of-Two Quantization: An Efficient Non-uniform Discretization for Neural Networks},
  booktitle = {International Conference on Learning Representations (ICLR)},
  year      = {2020},
}

@misc{lai_2018_cmsisnn,
  author        = {Lai, Liangzhen and Suda, Naveen and Chandra, Vikas},
  title         = {{CMSIS-NN}: Efficient Neural Network Kernels for {Arm Cortex-M} {CPUs}},
  year          = {2018},
  eprint        = {1801.06601},
  archivePrefix = {arXiv},
}

@inproceedings{david_2021_tflm,
  author    = {David, Robert and Duke, Jared and Jain, Advait and Janapa Reddi, Vijay and Jeffries, Nat and Li, Jian and Kreeger, Nick and Nappier, Ian and Natraj, Meghna and Wang, Tiezhen and Warden, Pete and Rhodes, Rocky},
  title     = {{TensorFlow Lite Micro}: Embedded Machine Learning for {TinyML} Systems},
  booktitle = {Proceedings of Machine Learning and Systems (MLSys)},
  volume    = {3},
  pages     = {800--811},
  year      = {2021},
}

@inproceedings{lin_2020_mcunet,
  author    = {Lin, Ji and Chen, Wei-Ming and Lin, Yujun and Cohn, John and Gan, Chuang and Han, Song},
  title     = {{MCUNet}: Tiny Deep Learning on {IoT} Devices},
  booktitle = {Advances in Neural Information Processing Systems (NeurIPS)},
  volume    = {33},
  pages     = {11711--11722},
  year      = {2020},
}

@misc{nagel_2021_whitepaper,
  author        = {Nagel, Markus and Fournarakis, Marios and Amjad, Rana Ali and Bondarenko, Yelysei and van Baalen, Mart and Blankevoort, Tijmen},
  title         = {A White Paper on Neural Network Quantization},
  year          = {2021},
  eprint        = {2106.08295},
  archivePrefix = {arXiv},
}

@inproceedings{rastegari_2016_xnor,
  author    = {Rastegari, Mohammad and Ordonez, Vicente and Redmon, Joseph and Farhadi, Ali},
  title     = {{XNOR-Net}: {ImageNet} Classification Using Binary Convolutional Neural Networks},
  booktitle = {European Conference on Computer Vision (ECCV)},
  pages     = {525--542},
  year      = {2016},
}

@inproceedings{dong_2019_hawq,
  author    = {Dong, Zhen and Yao, Zhewei and Gholami, Amir and Mahoney, Michael W. and Keutzer, Kurt},
  title     = {{HAWQ}: Hessian {AWare} Quantization of Neural Networks with Mixed-Precision},
  booktitle = {Proceedings of the IEEE/CVF International Conference on Computer Vision (ICCV)},
  pages     = {293--302},
  year      = {2019},
}

@article{vandersteegen_2021_int4shift,
  author  = {Vandersteegen, Maarten and Van Beeck, Kristof and Goedem{\'e}, Toon},
  title   = {Integer-Only {CNNs} with 4 Bit Weights and Bit-Shift Quantization Scales at Full-Precision Accuracy},
  journal = {Electronics},
  volume  = {10},
  number  = {22},
  pages   = {2823},
  year    = {2021},
  doi     = {10.3390/electronics10222823},
}
\bibliographystyle{mlsys2025}

\clearpage
\appendix

\section{Hardware-Friendly Quantization Properties and Their MAX7800x Mandate}
\label{app:hwfq}

\textbf{Uniform quantization.}
A uniform quantizer $Q(x) = \lfloor x/\Delta \rceil + Z$ uses a single step size $\Delta$ and zero-point $Z$ across the range, so multiply-accumulate operations need no lookup tables or branching.
The accelerator stores weights, biases, and data as fixed-point Q7 integers, which implies a uniform step; there is no documented non-uniform or lookup-table dequantization path.

\textbf{Symmetric quantization.}
Fixing $Z=0$ gives $Q(x) = \lfloor x/\Delta \rceil$ and removes the zero-point cross terms from every product.
The accelerator's accumulate-and-shift datapath assumes zero-centered operands.
The documented number format is signed two's complement with no offset term, so a nonzero zero-point would need an offset correction the datapath does not describe (C1).

\textbf{Power-of-two scaling.}
With $\Delta = 2^{s}/2^{n-1}$ for integer $s$, $Q(x) = \lfloor x\cdot 2^{n-1}/2^{s} \rceil$, and every rescaling becomes a shift.
On MAX7800x this is not a preference.
The per-layer output shift can only multiply by powers of two, so a non-power-of-two scale cannot be realized without a general multiplier on the output path (C2).

\textbf{Per-tensor quantization.}
Per-channel scales improve accuracy~\citep{habi_2021_hptq, li2021brecq, gordon_2024_eptq}, but they require a scale per output channel at inference.
The accelerator's output shift is configured per layer, so per-channel scaling would require per-channel shifts or a runtime multiply, neither of which is documented for this hardware generation (C3).

\section{Additional Experimental Details}
\label{app:exp}

\subsection{Hyperparameters}
\label{app:hyp}
\textbf{MobileNetV2.}
We use SGD with momentum 0.9 and weight decay $4\times10^{-5}$ for 200 epochs, with QAT starting at epoch 150.
Runs initialized from a pretrained checkpoint resume at epoch 150.
The learning rate follows cosine annealing from an initial value of 0.05.

\textbf{ResNet-18.}
We use SGD with momentum 0.9 and weight decay $10^{-4}$ for 120 epochs, with QAT starting at epoch 90.
Runs initialized from a pretrained checkpoint resume at epoch 90.
The learning rate follows MultiStepLR with milestones at epochs 30, 60, 90, and 105, and $\gamma=0.1$.

\textbf{Scale estimation.}
Activation statistics are collected as 2048-bin histograms in one forward pass over the training set, without backpropagation, immediately before QAT starts. Activation outliers are removed with a z-score threshold of 8.0. The activation-threshold search evaluates five power-of-two candidates, the non-clipping threshold and its first four halvings. The weight/bias range bound uses the 0.999 quantile for 8-bit layers and the 0.9 quantile for 4-bit layers. Total per-layer shifts are clamped to $[-15,15]$.

\subsection{Hardware Implementation Details}
\label{app:hwimpl}
To fit the memory and architectural limits of MAX78002, we made the following changes for deployment:
\begin{itemize}
\item Input resolution was set to 128$\times$128.
\item MobileNetV2 used width multiplier 0.75, and ResNet-18 used width multiplier 0.5.
\item Layers wider than 1024 channels were capped at 1024 channels, the hardware's per-layer channel limit.
\item Each stride-2 \texttt{conv2d} was replaced by 2$\times$2 max-pooling followed by a stride-1 \texttt{conv2d}.
\item Biases were quantized to 8 bits, as the hardware requires (C4). The accuracy comparisons in Tables~\ref{tab:w8a8} and~\ref{tab:w4a8} instead use 32-bit biases for all methods, matching the published baselines.
\end{itemize}

\subsection{Compute Resources}
All training ran on two NVIDIA RTX 4090 GPUs with 24~GB each.
Runs resumed from a pretrained checkpoint took approximately 12 hours, and runs from scratch took approximately 48 hours.

\subsection{Dataset License}
We use the ILSVRC-2012 subset of ImageNet under its non-commercial research license (\url{https://www.image-net.org/download.php}).

\end{document}